\documentclass{article}

\usepackage[preprint]{neurips_2024}
\usepackage[utf8]{inputenc} % allow utf-8 input
\usepackage[T1]{fontenc}    % use 8-bit T1 fonts
\usepackage{hyperref}       % hyperlinks
\usepackage{url}            % simple URL typesetting
\usepackage{booktabs}       % professional-quality tables
\usepackage{amsfonts}       % blackboard math symbols
\usepackage{nicefrac}       % compact symbols for 1/2, etc.
\usepackage{microtype}      % microtypography
\usepackage{xcolor}         % colors

\usepackage{graphicx}
\usepackage{amsmath}
\usepackage{amsthm}
\usepackage{algorithm}
\usepackage{algorithmic}
\usepackage{amssymb}

\usepackage{multirow}
\usepackage{authblk}
\usepackage{color}

\title{STG-Mamba: Spatial-Temporal Graph Learning via Selective State Space Model}

\author{
  Lincan Li, Hanchen Wang, Wenjie Zhang\thanks{Corresponding author.}\\
  University of New South Wales \\
  Sydney, Australia \\
  \texttt{lincan.li@unsw.edu.au, hanchen.wang@unsw.edu.au,} \\
  \texttt{wenjie.zhang@unsw.edu.au}\\
}

\begin{document}

\maketitle

\begin{abstract}
%Spatial-Temporal Graph (STG) data is characterized as dynamic, heterogenous, and non-stationary, leading to the continuous challenge of spatial-temporal graph learning. In the past few years, various GNN-based methods have been proposed to solely focus on mimicking the relationships among node individuals of the STG network, ignoring the significance of modeling the intrinsic features that exist in STG system over time. In contrast, state space models (SSMs) present a new approach which treat STG Network as a system, and meticulously explore the STG system's dynamic state evolution across temporal dimension. In this work, we introduce Spatial-Temporal Graph Mamba (STG-Mamba) as the first exploration of enhancing Long-Term Sequential Forecasting (LTSF) capability in STG learning tasks by treating STG Network as a system, and integrating a Graph State Space Module (GSSM) within the STG-Mamba framework. STG-Mamba is formulated as an Encoder-Decoder architecture, which is composed of N stacked Graph State Space Modules in both encoder and decoder, for efficient sequential data modeling. Extensive empirical studies are conducted on two benchmark STG forecasting datasets, demonstrating the performance superiority and computational efficiency of our STG-Mamba. It not only surpasses existing state-of-the-art methods in terms of short-term and long-range STG forecasting performance, but also effectively alleviate the computational bottleneck of LTSF task in reducing the computational cost of FLOPs and GPU memory consumption.
Spatial-Temporal Graph (STG) data is characterized as dynamic, heterogenous, and non-stationary, leading to the continuous challenge of spatial-temporal graph learning. In the past few years, various GNN-based methods have been proposed to solely focus on mimicking the relationships among node individuals of the STG network, ignoring the significance of modeling the intrinsic features that exist in STG system over time. In contrast, modern Selective State Space Models (SSSMs) present a new approach which treat STG Network as a system, and meticulously explore the STG system's dynamic state evolution across temporal dimension. In this work, we introduce Spatial-Temporal Graph Mamba (STG-Mamba) as the first exploration of leveraging the powerful selective state space models for STG learning by treating STG Network as a system, and employing the Spatial-Temporal Selective State Space Module (ST-S3M) to precisely focus on the selected STG latent features. Furthermore, to strengthen GNN's ability of modeling STG data under the setting of selective state space models, we propose Kalman Filtering Graph Neural Networks (KFGN) for dynamically integrate and upgrade the STG embeddings from different temporal granularities through a learnable Kalman Filtering statistical theory-based approach. Extensive empirical studies are conducted on three benchmark STG forecasting datasets, demonstrating the performance superiority and computational efficiency of STG-Mamba. It not only surpasses existing state-of-the-art methods in terms of STG forecasting performance, but also effectively alleviate the computational bottleneck of large-scale graph networks in reducing the computational cost of FLOPs and test inference time. The implementation code is available at: \url{https://github.com/LincanLi98/STG-Mamba}.
\end{abstract}
%\url{https://github.com/LincanLi98/STG-Mamba}.
%\url{https://anonymous.4open.science/r/STG-Mamba-81DB/}.

\section{Introduction}
Spatial-temporal graph data is a kind of Non-Euclidean data which widely exists in our daily life, such as urban traffic network, metro system InFlow/OutFlow, social networks, regional energy load, weather observations, etc. Owning to the dynamic, heterogenous, and non-stationary nature of STG data, accurate and efficient STG forecasting has long been a challenging task. 

Lately, with the popularization of Mamba~\cite{gu2023mamba,wang2024graph,liu2024swin}, Modern Selective State Space Models (SSSM) have aroused considerable interest among researchers from computer vision and Natural Language Processing. SSSM is a variant of the State Space Models, which originated from Control Science and Engineering fields~\cite{lee1994state,friedland2012control}. State Space Models provide a professional framework to describe a physical system's dynamics state evolution through sets of input, output, and state variables related by first-order differential or difference equations. The approach allows for a compact way to model and analyze systems with multiple inputs and outputs (MIMO)~\cite{aoki2013state}.

STG learning can be treated as a complex process of understanding and forecasting the evolution of STG networks over time, which highly resembles the state space transition process where each state encapsulates the current configuration of a system and transitions represent changes over time. Deep learning-based SSSMs bring new horizons to STG learning tasks. However, great challenges pose in accurately and effectively adopting SSSMs architectures for STG modeling.

Motivated by the excellent long-term contextual modeling ability and low computational overhead of SSSMs, we propose Spatial-Temporal Graph Mamba (STG-Mamba) as the first deep learning-based SSSM for effective data-centric STG learning. The main contributions of this work are:
%The main contributions of this work are summarized in the following:
\begin{itemize}
\item We make the first attempt to adapting SSSMs for STG learning tasks. A simple yet elegant way is employed to extend SSSMs to handle STG data. Specifically, we formulate the framework following the Stacked Residual Encoder Fashion, which takes $N$ stacked Graph Selective State Space Block (GS3B) as basic module, with the proposed Kalman Filtering Graph Neural Network (KFGN), the Spatial-Temporal Selective State Space Module (ST-S3M), and a simultaneous STG Feed-Forward connection to serialize and coordinate different internal modules. %{Emphasize here again the merits of KFGN}
\item A novel Spatial-Temporal Selective State Space Module (ST-S3M) is proposed for STG network pioneering integration with selective state space models, which performs input-dependent adaptive spatial-temporal graph feature selection. The Graph Selective Scan algorithm within ST-S3M simultaneously receives graph information from KFGN through the Feed-Forward connection, the Feed-Forward graph information is then employed to assist more effective updating of the state transition matrix and control matrix. 
%ST-S3M is capable of accurately modeling the dynamic evolution of STG system over time, making it highly fitted for STG learning.
\item We introduce KFGN as the specialized adaptive spatiotemporal graph generation and upgrading method, which smoothly fits in within the SSSM-based context. In KFGN, DynamicFilter-GNN serves as the module which generating input-specific dynamic graph structures. The KF-Upgrading mechanism models STG inputs from different temporal granularities as parallel streams, and the output embeddings are integrated through the statistical theory learning of Kalman Filtering for optimization. 
%\item We introduce Kalman Filtering Graph Neural Network (KFGN) as the specialized adaptive spatiotemporal graph generation and upgrading method, which smoothly fits in within the SSSM-based context, ensuring the synchronous updates between graph structure and STG systems's current state.   
\item Extensive evaluations are carried on three open-sourced benchmark STG dataset. Results demonstrate that our STG-Mamba not only exceed other benchmark methods in terms of STG prediction performance, but also achieve $O(n)$ computational complexity, remarkably reducing the computational overhead compared with Transformers-based methods.
\end{itemize}

\section{Preliminaries} \label{sec2}
%\noindent \textbf{Spatial-Temporal Graph (STG).} STG is a kind of non-Euclidean structured data, which extensively existed in daily-life, business, and society. For a given graph network, if the node individuals have certain spatial connectivity at any given time step $\tau$, and the node features and/or graph structures are dynamically evolving over temporal axis, it could be defined as an STG network. Formally, we use $\mathbb{G}^{ST}=(V^{ST},E^{ST},A^{ST})$ to represent a given STG network. Here, $V^{ST}$ denotes the set of graph vertices (nodes) within the network, $E^{ST}$ denotes the edges of the graph network. $A^{ST}$ is the graph adjacency matrix, which indicates the calculated proximity between node individuals. 

\noindent \textbf{Spatial-Temporal Graph System (STG system).} We for the first time define a network which consists of STG data as a Spatial-Temporal Graph System. In theory foundation, a system is defined as the representation of a physical process, comprising state variables that describe the system's current condition, input variables influencing the system's state, and output variables reflecting the system's response. For STG System, this framework is adapted to accommodate spatial dependencies and temporal evolution, structuring the system with nodes representing spatial entities, edges denoting spatial interactions, and state variables evolving over time to capture the dynamics of STG data.

\noindent \textbf{State Space Models (SSMs).} Deep Learning-based SSMs are a lately invented class of sequential models that are closely related to RNN architectures and classical state space models. They are characterized by a particular continuous system model that maps a multi-dimensional input sequence to a corresponding output sequence through an implicit latent state representation. SSMs are defined by four parameters $(\bold{A}, \bold{B}, \bold{C}, \bold{D})$ that specify how the input (control signal) and the current state determine the next state and output. This framework allows for efficient sequential modeling by enabling both linear and non-linear computations. As a variant of SSMs, SSSMs emphasize how to build the selection mechanism upon SSMs, which is highly resemble to the core idea of attention mechanism, making it a competitive counterpart of Transformer architecture.

%Formally, SSMs are expressed through state equations $x_{t+1} = f(x_t, u_t)$ and observation equations $y_t = g(x_t, u_t)$, where $x_t$ denotes the state at time $t$, $u_t$ represents the control input, and $y_t$ is the observed output. The state transition function $f(\cdot)$ describes how the current state $x_t$ evolves to the next state $x_{t+1}$ under the influence of the control input $u_t$. $f(\cdot)$ captures the dynamics of the system, accounting for how internal and external factors modify the system's state over time. The observation function $g(\cdot)$ is used for mapping the current state $x_t$ and control input $u_t$ to the observed output $y_t$. This function defines how the underlying state of the system is reflected in the measurements or observations collected.

%As a variant of SSMs, SSSMs emphasize how to build the selection mechanism upon SSMs, which is highly resemble to the core idea of attention mechanism, making it a competitive counterpart of Transformer architecture.

\noindent \textbf{STG Forecasting based on SSSMs.}
Employing SSSMs for spatial-temporal graph forecasting, the problem can be formulated as dynamically identifying and utilizing relevant portions of historical spatiotemporal data and graph structures to predict the future states of a STG system.

Given spatiotemporal graph $\mathbb{G}^{ST}=(V^{ST},E^{ST},A^{ST})$ and historical spatial-temporal sequential data $\rm{X}_{t-p+1:t}=\{X_{t-p+1},X_{t-p+2},...,X_{t}\}$, the objective is to utilize SSSMs to forecast future STG system's states $\rm{\hat{X}_{t+1:t+k}}=\{\hat{X}_{t+1},...,\hat{X}_{t+k}\}$. This process is achieved by learning a mapping function $F_{SSSM}(\cdot)$ that dynamically selects relevant state transitions and interactions for prediction:
\begin{equation}
\setlength{\abovedisplayskip}{3pt}
\setlength{\belowdisplayskip}{3pt}
F_{SSSM}(\rm{X}_{t-p+1:t} ; \mathbb{G}^{ST})= \rm{\hat{X}_{t+1:t+k}}
\label{eq:eq1}
\end{equation}

\section{Methodology} \label{sec3}

In this section, we start with introducing the model architecture of STG-Mamba and the formulation of STG Selective State Space Models in Section~\ref{sec3-1} and Appendix~\ref{appendD} as well. Following that, we elaborate the key contributions of this work, including the Kalman Filtering Graph Neural Networks (Section~\ref{sec3-2}) which specifically designed for SSSMs to enable adaptive statistical-based graph learning under contexts with noisy data and uncertainty; and the Spatial-Temporal Selective State Space Module (ST-S3M) which serves as the state space selection mechanism for spatial-temporal graph learning (Section~\ref{sec3-3}). Finally, a detailed computational efficiency analysis of STG-Mamba in presented in Appendix~\ref{appendB} because of limited space.

\subsection{Architecture of STG Selective State Space Models}\label{sec3-1}

\begin{figure*}[htb]
  \centering
  \includegraphics[width=0.96\columnwidth]{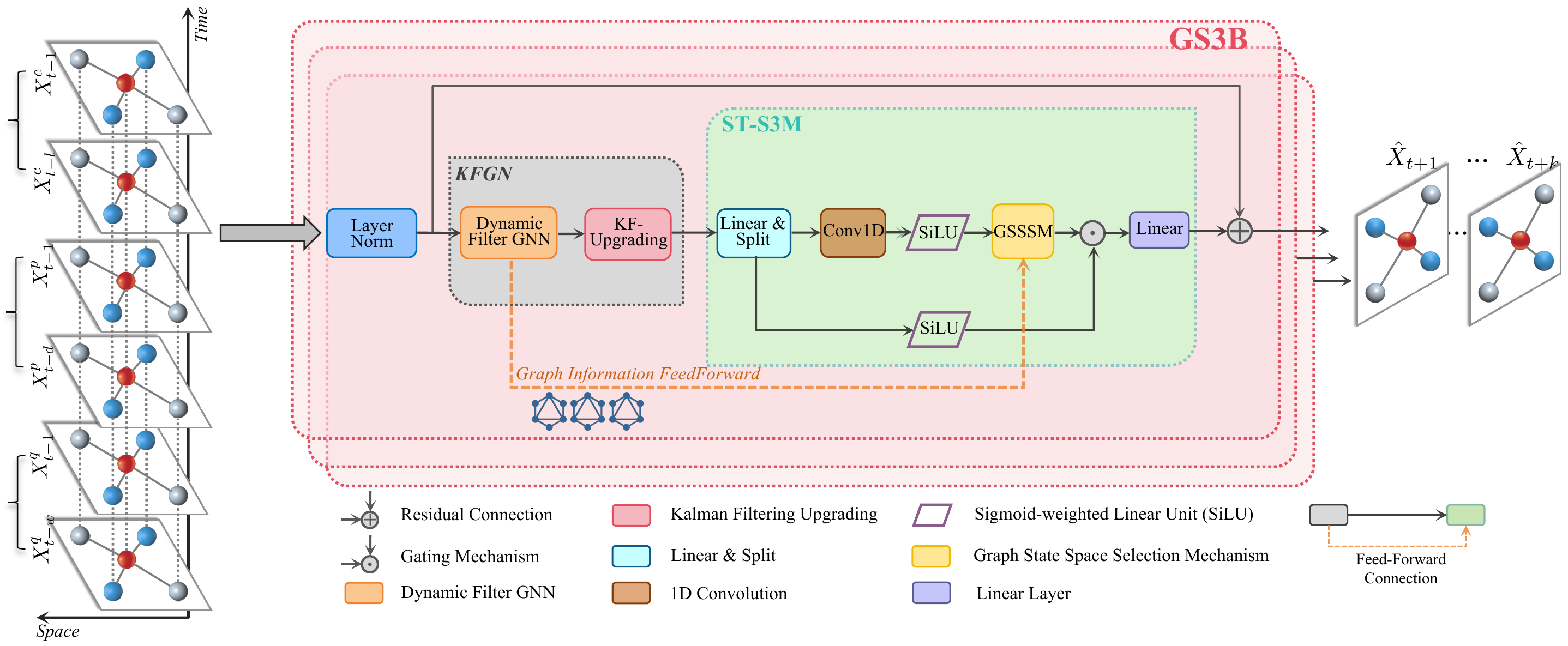}
  \caption{The comprehensive illustration of STG-Mamba's Model Architecture.}
  \label{fig1}
\end{figure*}

Figure~\ref{fig1} illustrates the comprehensive architecture of the proposed STG-Mamba. Specifically, we formulate the overall architecture following the Residual Encoder fashion for efficient sequential data modeling and prediction. In STG-Mamba, we leverage the Graph Selective State Space Block (GS3B) as the basic Encoder module, and repeat it $N$ times. GS3B consists of several networks and operations, including Layer Norm, Kalman Filter Graph Neural Networks (KFGN) which consists of DynamicFilter-GNN followed by Kalman Filter Upgrading (KF-Upgrading), Spatial-Temporal Selective State Space Module (ST-S3M) which consists of Linear \& Split, Conv1D, SiLU, Grapg State Space Selection Mechanism (GSSSM), Element-Wise Concatenation, Graph Information Feed-Forward, as well as the Residual Connection. The Graph Information Feed-Forward structure here is specifically designed for coordinating different modules' information transmission and updating, ensuring the latest STG information is available for every module.

\subsection{Kalman Filtering Graph Neural Networks}\label{sec3-2}
The motivation for employing Kalman Filtering-based methods in constructing GNNs stems from the need to enhance the reliability and accuracy of spatial-temporal forecasting. STG big data (i.e., traffic sensor records, weather station records, etc) often contains inherent biases and noises that existing methods typically overlook. By integrating Kalman Filtering-based optimization and upgrading, KFGN effectively addresses these inaccuracies by dynamically weighing the reliability of data streams from different temporal granularity, optimizing the fusion of these data streams based on the estimated variances. This approach not only corrects dataset inherent errors but also significantly improves the model's capacity to capture the complex inter-dependencies within STG patterns. 

\begin{figure*}[!htb]
  \centering
  \includegraphics[width=0.98\columnwidth]{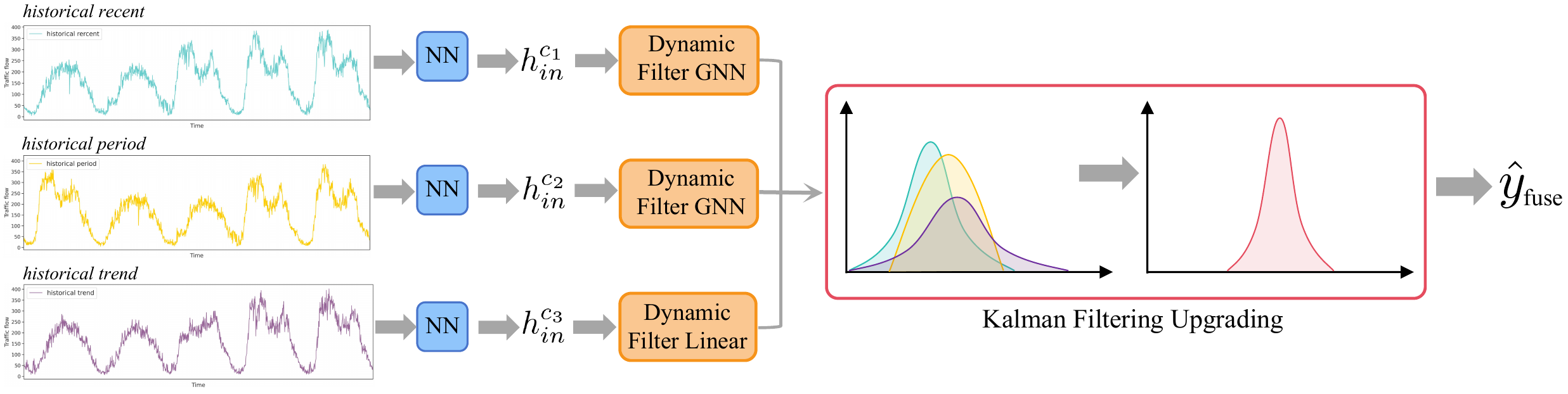}
  \caption{Kalman Filtering Graph Neural Networks.}
  \label{fig2}
\end{figure*}

As depicted in Figure~\ref{fig2}, the KFGN pipeline consists of two key steps. In the first step, embeddings generated from the model inputs of different temporal granularities (i.e., recent steps/periodic steps/trend steps) are sent into a DynamicFilter-GNN module for processing. Then, in the second step, the outputs of DynamicFilter-GNN module are integrated and optimized through the Kalman Filtering Upgrading module.
In the initial stage of DynamicFilter-GNN, a learnable Base Filter parameter matrix of size $\mathbb{R}^{\text{in\_fea}\times \text{in\_fea}}$ is defined. It's employed to transform the graph adjacency matrix, thereby dynamically adjusting the connection degrees between nodes.
Following that, we uniformly initialize the weights and biases, with an initialization standard deviation $\text{stdv}$ calculated as the reciprocal of the square root of the number of input features: $\text{stdv} = \frac{1}{\sqrt{\text{in\_fea}}}$. This uniform initialization is critical for ensuring the module starts from a neutral point.

Subsequently, we transformer the Base Filter via a linear transformation layer, and combine it with the original adjacency matrix $A_{\text{ini}}^{c_i}$, resulting in a dynamically adjusted adjacency matrix $A_{\text{dyn}}^{c_i}$. Let $c_{i,i = \{1,2,3\}} \in \{r,p,q\}$ represents the input data from different temporal granularities, (i.e., $c_1=r$ denotes historical recent, $c_2=p$ denotes historical period, $c_3=q$ denotes historical trend). 
Leveraging $A_{\text{dyn}}^{c_i}$, input embedding $h_{in}^{c_i}$, weights $W_{DF}^{c_i}$ and bias $b_{DF}^{c_i}$, the input embedding undergoes a graph convolution:
\begin{equation}
\setlength{\abovedisplayskip}{3pt}
\setlength{\belowdisplayskip}{3pt}
h_{DF}^{c_i}=h_{in}^{c_i} \cdot (A_{\text{dyn}}^{c_i}\cdot W_{DF}^{c_i})+b_{DF}^{c_i}
\label{eq:eq5}
\end{equation}
This design enables the model to dynamically adjust the strengths of connections between nodes within the graph based on the learned adjustments.

Following DynamicFilter-GNN, the next step is Kalman Filtering Upgrading. Since the input data are subjected to the same dataset but within different temporal granularities, they are assumed to follow Gaussian Distributions:
\begin{equation}
\setlength{\abovedisplayskip}{3pt}
\setlength{\belowdisplayskip}{3pt}
y_q(x;\mu_q,\sigma_q) \triangleq \frac{\exp^{-\frac{(x-\mu_q)^2}{2 \sigma_q^2}}}{\sqrt{2 \pi \sigma_q^2}}; 
y_p(x;\mu_p,\sigma_p) \triangleq \frac{\exp^{-\frac{(x-\mu_p)^2}{2 \sigma_p^2}}}{\sqrt{2 \pi \sigma_p^2}};
y_r(x;\mu_r,\sigma_r) \triangleq \frac{\exp^{-\frac{(x-\mu_r)^2}{2 \sigma_r^2}}}{\sqrt{2 \pi \sigma_r^2}};
\label{eq:eq6}
\end{equation}
where subscript $q=c_3, p=c_2, r=c_1$ denote historical trend/period/recent dataset, respectively.

We employ Kalman Filtering method to derive accurate information from the multi-granularity observation sets. Specifically, their individual probability distribution functions can be integrated together by multiplication:
\begin{align}
\setlength{\abovedisplayskip}{3pt}
\setlength{\belowdisplayskip}{3pt}
y_{fuse}(x;\mu_q,\sigma_q,\mu_p,\sigma_p,\mu_r,\sigma_r) &= \frac{\exp^{-\frac{(x-\mu_q)^2}{2 \sigma_q^2}}}{\sqrt{2 \pi \sigma_q^2}} \times \frac{\exp^{-\frac{(x-\mu_p)^2}{2 \sigma_p^2}}}{\sqrt{2 \pi \sigma_p^2}} \times \frac{\exp^{-\frac{(x-\mu_r)^2}{2 \sigma_r^2}}}{\sqrt{2 \pi \sigma_r^2}} \nonumber
\\
&= \frac{1}{(2\pi)^{3/2} \sqrt{\sigma_q^2\sigma_p^2\sigma_r^2}} \exp^{-\left(\frac{(x-\mu_q)^2}{2 \sigma_q^2} + \frac{(x-\mu_p)^2}{2 \sigma_p^2} + \frac{(x-\mu_r)^2}{2 \sigma_r^2}\right)}
\label{eq:eq7}
\end{align}

By reorganizing Eq.~\ref{eq:eq7} into a simplified version, we have:
\begin{equation}
\setlength{\abovedisplayskip}{3pt}
\setlength{\belowdisplayskip}{3pt}
y_{fuse}(x;\mu_{fuse},\sigma_{fuse})=\frac{1}{\sqrt{2\pi \sigma_{fuse}^2}} \exp^{-\frac{(x-\mu_{fuse})^2}{2\sigma_{fuse}^2}}
\label{eq:eq8}
\end{equation}
\noindent where $\mu_{fuse} = \frac{\mu_q / \sigma_q^2 + \mu_p / \sigma_p^2 + \mu_r / \sigma_r^2}{1 / \sigma_q^2 + 1 / \sigma_p^2 + 1 / \sigma_r^2}$ and $\sigma_{fuse}^2 = \frac{1}{1 / \sigma_q^2 + 1 / \sigma_p^2 + 1 / \sigma_r^2}$.

To simplify the representation of $\mu_{fuse}$ and $\sigma_{fuse}^2$, we hereby introduce parameters $\omega_q=\frac{1}{\sigma_q^2}, \omega_p=\frac{1}{\sigma_p^2}, \omega_r=\frac{1}{\sigma_r^2}$. Then, $\mu_{fuse}$ and $\sigma_{fuse}^2$ can be re-written as:
\begin{equation}
\setlength{\abovedisplayskip}{3pt}
\setlength{\belowdisplayskip}{3pt}
\mu_{fuse} = \frac{\mu_q \omega_q + \mu_p \omega_p + \mu_r \omega_r}{\omega_q + \omega_p + \omega_r};
\\
\sigma_{fuse}^2 = \frac{1}{\omega_q + \omega_p + \omega_r}
\label{eq:eq9}
\end{equation}

This means that observations from different branches can be effectively integrated using weighted sum, and the weight is a combination of variances:
\begin{eqnarray}
\setlength{\abovedisplayskip}{3pt}
\setlength{\belowdisplayskip}{3pt}
\mu_{fuse} &=&\mu_q (\frac{\omega_q}{\omega_q + \omega_p + \omega_r})+\mu_p (\frac{\omega_p}{\omega_q + \omega_p + \omega_r})+\mu_r(\frac{\omega_r}{\omega_q + \omega_p + \omega_r}) \nonumber
\\
&\;& \Downarrow \label{eq:eq10}
\\
y_{fuse} &=& y_q(\frac{\omega_q}{\omega_q + \omega_p + \omega_r})+y_p (\frac{\omega_p}{\omega_q + \omega_p + \omega_r})+ y_r (\frac{\omega_r}{\omega_q + \omega_p + \omega_r}) \nonumber
\end{eqnarray}
Directly calculating the variance of an observation set requires the complete data, which is computationally expensive. Therefore, we estimate the variance by computing the variance distribution of each training samples:
\begin{equation}
\setlength{\abovedisplayskip}{3pt}
\setlength{\belowdisplayskip}{3pt}
\mathbb{E}[\sigma_{\{q,p,r\}}^2]=\frac{1}{N_m} \sum_{i} \frac{(S_i-\overline{S})^2}{L}
\label{eq:eq11}
\end{equation}
where $L$ is the length of each sample sequence, and $N_m$ is the number of data samples. $S_i$ denotes the $i$-th observation, and $\overline{S}$ denotes the average value of all observed samples. To further improve the integration and upgrading, we add two learnable weight params $\epsilon, \varphi$ to balance different observation branches. To this end, based on the formulation of $y_{fuse}$ in Eq.~\ref{eq:eq10}, the output of Kalman Filtering Upgrading module is:
\begin{equation}
\setlength{\abovedisplayskip}{3pt}
\setlength{\belowdisplayskip}{3pt}
\tilde{y}_{fuse}=\frac{\epsilon \cdot (\hat{y}_q \omega_q)+\varphi \cdot (\hat{y}_p \omega_p)+\hat{y}_r \omega_r}{\omega_q+\omega_p+\omega_r}
\label{eq:eq12}
\end{equation}
\noindent where $\omega_q=\frac{1}{\sigma_q^2}, \omega_p=\frac{1}{\sigma_p^2}, \omega_r=\frac{1}{\sigma_r^2}$, $\epsilon$ and $\varphi$ are trainable weights. Finally, for the convenience of neural network training and ensure the scalability, we remove the constant denominator $(\omega_q+\omega_p+\omega_r)$ in Eq.~\ref{eq:eq12}. As such, the final output of Kalman Filtering Upgrading is defined as:
\begin{equation}
\setlength{\abovedisplayskip}{3pt}
\setlength{\belowdisplayskip}{3pt}
\hat{y}_{fuse}=\epsilon \cdot (\hat{y}_q \omega_q)+\varphi \cdot (\hat{y}_p \omega_p)+\hat{y}_r \omega_r
\label{eq:eq13}
\end{equation}
Here, $\hat{y}_{fuse}$ is the final output of KF-Upgrading module. Note that our approach is an optimized version of the classic Kalman Filtering theory, which is specifically designed for deep learning-based methodology, ensuring both computationally efficiency, dynamic hierarchical STG feature fusion, and improved accuracy.

\subsection{ST-S3M: Spatial-Temporal Selective State Space Module}\label{sec3-3} 
ST-S3M plays the significant role of adaptive STG feature selection, serving as the counterpart of Attention mechanism. The architecture of ST-S3M is shown in Figure~\ref{fig1}. After being processed by KFGN module for dynamic spatiotemporal dependency modeling and statistical-based integration \& upgrading. The generated embedding $\hat{h}_{fuse}$ is sent to ST-S3M for input-specific dynamic feature selection. In ST-S3M, the input STG embedding passes through a Linear Layer, followed by a splitting operation. Let $b$ denotes batch size, $l$ denotes sequence length, $d_{model}$ be the feature dimension, $d_{inner}$ be the model inner feature dimension, we have:
\begin{align}
\setlength{\abovedisplayskip}{3pt}
\setlength{\belowdisplayskip}{3pt}
h_{\text{main-res}}&=W_{in} \hat{h}_{fuse}+b_{in}\nonumber
\\
(h_{\text{main}},res)&=\text{split}(h_{\text{main-res}})
\label{eq:eq14}
\end{align}
where $\hat{h}_{fuse}\in \mathbb{R}^{b\times l\times d_{model}}$, $W_{in} \in \mathbb{R}^{2 d_{inner}\times d_{model}}$, $b_{in} \in \mathbb{R}^{2 d_{inner}}$, $h_{\text{main-res}} \in \mathbb{R}^{b\times l \times 2d_{inner}}$, $h_{\text{main}}, res \in\mathbb{R}^{b\times l \times d_{inner}}$ 

Then, $h_{\text{main}}$ flows into a 1D convolution layer, followed by SiLU activation function:
\begin{equation}
\setlength{\abovedisplayskip}{3pt}
\setlength{\belowdisplayskip}{3pt}
h_{\text{main}}^{'}=\text{SiLU}(\text{Conv1D}(h_{\text{main}}))
\label{eq:eq15}
\end{equation}

\noindent where $h_{\text{main}}^{'} \in \mathbb{R}^{b\times l \times d_{inner}}$. The output of SiLU is sent to graph state space selection mechanism:
\begin{equation}
\setlength{\abovedisplayskip}{3pt}
\setlength{\belowdisplayskip}{3pt}
h_{\text{sssm}}=\text{GSSSM}(h_{main}^{'})
\label{eq:eq16}
\end{equation}
where $h_{\text{sssm}} \in \mathbb{R}^{b\times l \times d_{inner}}$. Meanwhile, the residual part $res$ is also processed by SiLU activation, and finally we employ element-wise dot product to fuse the main STG embedding and its residual: $h_{\text{sssm}}\odot \text{SiLU}(res)$. The fused result is transformed through a Linear projection:
\begin{equation}
\setlength{\abovedisplayskip}{3pt}
\setlength{\belowdisplayskip}{3pt}
h_{out}=W_{out} (h_{\text{sssm}}\odot \text{SiLU}(res))+b_{out}
\label{eq:eq17}
\end{equation}

\noindent where $h_{out} \in \mathbb{R}^{b \times l \times d_{model}}$ is the final output of ST-S3M. $W_{out} \in \mathbb{R}^{d_{model}\times d_{inner}}$, $b_{out} \in \mathbb{R}^{d_{model}}$.

The GSSSM in ST-S3M plays a main role in adaptive spatiotemporal feature selection. We detail the parameter computation and update process of the Graph State Space Selection Mechanism in Algorithm~\ref{alg:alg2}. Graph Selective Scan Algorithm (Algorithm~\ref{alg:alg1}) which receives graph information feed-forward is the most significant step in the state space selection process. We detail it in the following.

\noindent \textbf{Graph Selective Scan Algorithm.} The algorithm is a further extension of the basic selective scan, which integrates dynamic graph information generated by KFGN into state-space selection \& update procedure, enhancing Mamba's capability in capturing STG dependencies. The key steps and modifications are as follows:

We start with obtaining the dimension of the input tensor $u \in \mathbb{R}^{(b,l,d_{in})}$, where $b$ is the batch size, $l$ denotes the sequence length, $d_{in}$ is the input features dim; and the second dim of $\bold{A}$ is denoted as $n$. 

Next, we highlight the main novelty of the graph selective scan algorithm--the \textbf{integrated upgrade of graph information Feed-Forward and param $\Delta^*$}, which corresponds to line 2-line 5 of Algorithm~\ref{alg:alg1}. We retrieve the dynamic graph adjacency matrix $\alpha_t$from the DynamicFilter-GNN: $\alpha_t=\text{DynamicFilter-GNN}.\text{get\_transformed\_adjacency()}$. In order to let the graph information to participate in the state space selection and update process, an intuitive and natural way is to fuse the  parameter $\Delta^*$, and let the fused parameter flow into the following calculation of $\text{delta}\bold{A}$ and $\text{delta}\bold{B}_u$. However, the dimension of $\alpha_t$ and $\Delta^*$ may not be identical. Thus, we initialize a padding matrix $\text{adj\_padded}=\mathbf{1}^{d_{\text{in}} \times d_{\text{in}}}$, then we fill the padding matrix with the graph information from $\alpha_t$ as: $\text{adj\_padded}[:\alpha_t.\text{size}(0),:\alpha_t.\text{size}(1)]=\alpha_t$. After the dimension adjustment, we therefore integrate $\Delta^*$ and $\text{adj\_padded}$ by performing matrix multiplication: $\Delta^{'}=\text{matmul}(\Delta^*,\text{adj\_padded})$. 

Following that, we Discretize the continuous parameters $\bold{A}$ and $\bold{B}$ as: %more effective updating of the state transition matrix and control matrix. 
\begin{align}
\setlength{\abovedisplayskip}{3pt}
\setlength{\belowdisplayskip}{3pt}
& \text{State Transition Matrix update: } \text{delta}\bold{A} = \exp(\text{einsum}(\Delta^{'},\bold{A})) \label{eq:eq18}
\\
& \text{Control Matrix update: } \text{delta}\bold{B}_u = \text{einsum}(\Delta^{'},\bold{B},u)\nonumber
\end{align}

\noindent where $\text{einsum}$ denotes the Einstein Summation Convention. $\text{delta}\bold{A}$ is the updated state transition matrix, $\text{delta}\bold{B}_u$ is the updated control matrix.

Then, Iterative State Update is executed on state $x \in \mathbb{R}^{b \times d_{in}\times n}$. For time step from $i$ to $l$, we have:
\begin{align}
\setlength{\abovedisplayskip}{3pt}
\setlength{\belowdisplayskip}{3pt}
x & \gets \text{delta}\bold{A}[:, i] \times x + \text{delta}\bold{B}_u[:, i] \label{eq:eq20}
\\
z & \gets \text{einsum}(x, \bold{C}[:, i, :]) \nonumber
\end{align}

\noindent where $z$ is the current output computed by einstein summation convention. $z$ is then appended to the output list $z_s$. The list of outputs $z_s$ is stacked to form the output tensor $z$, and finally we add the direct gain $\bold{D}$ to the final output as: $z \gets z + u \times \bold{D}$.

The Graph Selective Scan algorithm offers several advantages including: enhanced spatiotemporal dependency modeling, adaptability to changing graph structures, and improved forecasting accuracy, making it particularly suitable for spatiotemporal graph learning tasks.

\section{Experiments} \label{sec4}

\subsection{Dataset Statistics and Baseline Methods} \label{sec4-1}
\noindent \textbf{Datasets.} We meticulously select three real-world STG datasets from California road network speed records, Hangzhou metro system entry/exit records, and weather station records across mainland China. Namely PeMS04~\cite{song2020spatial}, HZMetro~\cite{9269513}, and KnowAir~\cite{wang2020pm2}, respectively. The detailed dataset statistics and descriptions are summarized in Table~\ref{tab1}.

%\noindent \textbf{Datasets.} For experimental evaluation, we meticulously select three real-world STG datasets from California road network speed records (PeMS04)~\cite{song2020spatial}, Hangzhou urban metro system entry/exit records (HZMetro)~\cite{9269513}, and weather station records across mainland China (KnowAir)~\cite{wang2020pm2}. Acquired from the Caltrans Performance Measurement System (PeMS), PeMS04 records the average traffic speed from a total number of 307 sensors distributed on California Highway system. The time range of PeMS04 is 1st/Jan/2018 - 28th/Feb/2018, with 5-minute time interval. HZMetro was created with the In/Out transaction records of 80 stations of Hangzhou Metro system from 1st Jan 2019 to 25th Jan 2019, which takes 15-minute as data collection time interval. KnowAir documents weather observations in every 3-hour, encompassing a dataset that spans from 2015 to 2018. Within the KnowAir dataset, there are three subsets. In this study, we exclusively use subset No. 3, covering the period from 1/Sep/2016 to 31/Jan/2017. The detailed dataset statistics and descriptions are summarized in Table~\ref{tab1}. 

\begin{table}[htb]
\centering
\footnotesize
\caption{Statistics and descriptions of the three datasets used for experimental evaluation.}
\begin{tabular}{c|c|c|c}
\hline
\textbf{Dataset}  & \textbf{PeMS04}  & \textbf{HZMetro}  & \textbf{KnowAir}             \\ \hline
City         & California, USA          & Hangzhou, China          & 184 main cities across China \\ %\hline
Data Type      & Network traffic speed    & Station InFlow/OutFlow   & Weather records              \\ %\hline
Total Nodes    & 307                      & 80                       & 184                          \\ %\hline
Time Interval  & 5-minute               & 15-minute               & 3-hour                       \\ %\hline
Time Range     & 1/Jan/2018 - 28/Feb/2018 & 1/Jan/2019 - 25/Jan/2019 & 1/Sep/2016 - 31/Jan/2017     \\ %\hline
Dataset Length & 16,992                   & 2,400                    & 1,224                        \\ \hline
\end{tabular}
\label{tab1}
\end{table}

\noindent \textbf{Baseline Methods.} For fair comparison, we consider both benchmark spatiotemporal graph neural network (STGNN)-based methods, and Transformer-based methods for STG learning.  

\textit{STGNN-based Methods:} (I) STGCN~\cite{yu2018spatio}, (II) STSGCN~\cite{song2020spatial}, (III) STG-NCDE~\cite{choi2022graph}, (IV) DDGCRN~\cite{weng2023decomposition}.

%(I) STGCN~\cite{yu2018spatio}: STGCN propose to capture the complex spatial-temporal dependencies within traffic network by graph neural networks, and design a fully convolutional approach for fast training.

%(II) STSGCN~\cite{song2020spatial}: STSGCN presents a novel model which efficiently handling complex localized spatial-temporal correlations and data heterogeneity through a synchronous mechanism. 

%(III) STG-NCDE~\cite{choi2022graph}: STG-NCDE leverages the concept of Neural Controlled Differential Equations (NCDEs) for spatial-temporal processing within a unified framework, achieving superior accuracy across six benchmark datasets.

%(IV) DDGCRN~\cite{weng2023decomposition}: DDGCRN dynamically captures the spatial-temporal features by separating the normal and abnormal signals and employing a novel graph convolutional recurrent network approach. 

\textit{Attention (Transformer)-based Methods:} (V) ASTGCN~\cite{guo2019attention}, (VI) ASTGNN~\cite{guo2021learning}, (VII) PDFormer~\cite{jiang2023pdformer}, (VIII) STAEformer~\cite{liu2023spatio}, (IX) MultiSPANS~\cite{zou2024multispans}.

%(V) ASTGCN~\cite{guo2019attention}: ASTGCN leverages an attention-based spatial-temporal graph convolutional network to accurately predict STG data by dynamically capturing complex spatiotemporal correlations and patterns.

%(VI) ASTGNN~\cite{guo2021learning}: Attention based Spatial-Temporal Graph Neural Network incorporates a unique trend-aware self-attention and dynamic graph convolution to address temporal dynamics, spatial correlations, and the inherent periodicity and heterogeneity of STG data.

%(VII) PDFormer~\cite{jiang2023pdformer}: PDFormer introduces a tailored Transformer model, adept at capturing dynamic spatial dependencies and addressing the time delay in traffic condition propagation with its unique spatial self-attention and traffic delay-aware feature transformation.

%(VIII) STAEformer~\cite{liu2023spatio}: STAEformer introduces a novel approach that enhances vanilla Transformer with a spatiotemporal adaptive embedding component, enabling it to capture complex STG patterns.

%(IX) MultiSPANS~\cite{zou2024multispans}: MultiSPANS leverages multi-filter convolution modules, Transformers, and structural entropy optimization to effectively model complex multi-range dependencies in STG forecasting.

\subsection{Implementation Settings} \label{sec4-2}
We split all the datasets with a ratio of 6:2:2 for Training/Validation/Testing sets along the time axis, respectively. Before starting model training session, all the data samples are normalized into range $[0,1]$ with MinMax normalization method. We set the batch size as 48 and employ AdamW as the Optimizer. The learning rate is initialized as $1e^{-4}$, with a weight decay of $1e^{-2}$. CosineAnnealingLR~\cite{loshchilov2016sgdr} is adopted as the learning rate scheduler, with a maximum of 50 iterations, and a minimum learning rate $1e^{-5}$. Model training epochs is set as 100. The number of GS3B Encoder Layer is set as $N=4$. We use the historical ground-truth data from the past 12 steps (for historical recent/period/trend scales) to forecast the future 12 time steps. Mean Squared Error (MSE) is employed as the Loss Function for optimization. All experiments are conducted under one NVIDIA A100 80GB GPU for neural computing acceleration. 
%And finally, we employ RMSE/MAE/MAPE/$std_{\text{MAE}}$ as the deterministic evaluation metrics, $\rm{\Delta_{\text{VAR}}}$ (Variance Accounted For) and $\rm{R^2}$ (Coefficient of Determination) as statistic-based evaluation metrics. The reported results in Experiment section are the average value of 10 runs.

\subsection{Evaluation Metrics} \label{sec4-3}
We employ RMSE/MAE/MAPE/$std_{\text{MAE}}$ as the \textbf{deterministic evaluation metrics}, all the indicators are smaller the better; $\rm{\Delta_{\text{VAR}}}$ (Change in Variance Accounted For) and $\rm{R^2}$ (Coefficient of Determination) as \textbf{statistic-based evaluation metric}, where $\rm{\Delta_{\text{VAR}}}$ is the smaller the better, $\rm{R^2}$ is the larger the better. The reported results in Experiment section are the average value of 10 runs. The detailed indicator calculation methods are provided in Appendix~\ref{appendB} because of limited space.

\subsection{Results Evaluation and Comparison}

%\subsubsection{Experimental Results Comparison on Three STG Forecasting Tasks}
A comprehensive evaluation between STG-Mamba and the proposed baseline methods is conducted, and we report the RMSE/MAE/MAPE/$std_{\text{MAE}}$ performance results in Table~\ref{tab2}. STG-Mamba consistently outperforms other benchmark methods, except for the MAPE criteria of PeMS04 (Flow). It can also be easily observed that Transformer-based methods which integrate GNNs inside the architecture always gain superior performance than sole GNN-based methods. For instance, PDFormer, STAEformer, and MultiSPANS always show better performance than STGCN, STSGCN, and STG-NCDE. The reason is that although GNN is an effective approach of modeling STG tasks, Transformer-based methods have stronger ability to capture local and long-term dynamic dependencies. Within the GNN-based methods, STG-NCDE and DDGCRN stand out for their competitive performance. STG-NCDE models continuous-time dynamics through neural controlled differential equations (NCDE), offering a more precise representation of spatiotemporal feature changes over time compared to other discrete GNN-based methods. 

As a new contender poised to challenge Transformer architecture, the selective state space mechanism of Mamba diverges from Attention by utilizing a continuous-time model that captures the dynamic evolution of spatial-temporal dependencies more naturally and efficiently. Unlike Attention, SSSM directly captures the temporal evolution of features, enabling it to scale more gracefully with the sequence length and complexity of STG data. Furthermore, we facilitate STG-Mamba with Kalman Filtering-based adaptive STG evolution, the elaborately designed graph selective scan algorithm, and the Feed-Forward connection, making it inherit the advantages of modern SSSM, and highly suitable for STG learning. 

%As a new contender poised to challenge Transformer architecture, Mamba introduces an innovative approach by replacing the foundational Attention mechanism with modern selective state space models (SSSM). SSSM diverges from Attention by utilizing a continuous-time model that captures the dynamic evolution of spatial-temporal dependencies more naturally and efficiently. Unlike Attention, which computes weights across all pairs of input-output positions and can be computationally intensive, SSSM directly captures the temporal evolution of features, enabling it to scale more gracefully with the sequence length and complexity of STG data. Furthermore, we facilitate STG-Mamba with Deep Kalman Filtering-based adaptive spatiotemporal graph evolution, and the elaborately designed ST-S3M, making it inherit the advantages of modern SSSM, and highly suitable for STG learning. 

\begin{table}[htb]
\centering
\scriptsize
\caption{Performance Eval and Comparison with Baselines on PeMS04/HZMetro/KnowAir Dataset.}
\label{tab2}
\resizebox{0.99\textwidth}{!}{
\begin{tabular}{c|cccc|cccc|cccc}
\hline
\multirow{2}{*}{\textbf{Model}} & \multicolumn{4}{c|}{\textbf{PeMS04 (Flow)}}                                                                                             & \multicolumn{4}{c|}{\textbf{HZMetro}}                                                                                                   & \multicolumn{4}{c}{\textbf{KnowAir}}                                                                                                   \\ \cline{2-13} 
                                & \multicolumn{1}{c|}{\textbf{RMSE}}  & \multicolumn{1}{c|}{\textbf{MAE}} & \multicolumn{1}{c|}{\textbf{MAPE}}  & \textbf{$std_{\text{MAE}}$} & \multicolumn{1}{c|}{\textbf{RMSE}}  & \multicolumn{1}{c|}{\textbf{MAE}}   & \multicolumn{1}{c|}{\textbf{MAPE}}  & \textbf{$std_{\text{MAE}}$} & \multicolumn{1}{c|}{\textbf{RMSE}}  & \multicolumn{1}{c|}{\textbf{MAE}} & \multicolumn{1}{c|}{\textbf{MAPE}} & \textbf{$std_{\text{MAE}}$} \\ \hline
STGCN                           & \multicolumn{1}{c|}{35.55}          & \multicolumn{1}{c|}{22.70}          & \multicolumn{1}{c|}{14.59}          & 1.2068               & \multicolumn{1}{c|}{34.85}          & \multicolumn{1}{c|}{21.33}          & \multicolumn{1}{c|}{13.47}          & 1.1835               & \multicolumn{1}{c|}{11.46}      & \multicolumn{1}{c|}{8.37}          & \multicolumn{1}{c|}{11.26}          & 0.4358               \\
STSGCN                          & \multicolumn{1}{c|}{33.65}          & \multicolumn{1}{c|}{21.19}          & \multicolumn{1}{c|}{13.90}          & 1.1820               & \multicolumn{1}{c|}{33.24}          & \multicolumn{1}{c|}{20.79}          & \multicolumn{1}{c|}{13.06}          & 1.1249               & \multicolumn{1}{c|}{11.22}      & \multicolumn{1}{c|}{8.15}          & \multicolumn{1}{c|}{10.89}          & 0.4134               \\
STG-NCDE                        & \multicolumn{1}{c|}{31.09}          & \multicolumn{1}{c|}{19.21}          & \multicolumn{1}{c|}{12.76}          & 1.0736               & \multicolumn{1}{c|}{32.91}          & \multicolumn{1}{c|}{20.75}          & \multicolumn{1}{c|}{12.88}          & 1.1097               & \multicolumn{1}{c|}{10.85}          & \multicolumn{1}{c|}{7.93}          & \multicolumn{1}{c|}{10.47}          & 0.3922               \\ 
DDGCRN                          & \multicolumn{1}{c|}{30.51}          & \multicolumn{1}{c|}{18.45}          & \multicolumn{1}{c|}{12.19}          & 1.0365               & \multicolumn{1}{c|}{31.69}          & \multicolumn{1}{c|}{19.52}          & \multicolumn{1}{c|}{12.45}          & 1.0586               & \multicolumn{1}{c|}{10.43}          & \multicolumn{1}{c|}{7.84}          & \multicolumn{1}{c|}{10.38}          & 0.3794               \\
ASTGCN                          & \multicolumn{1}{c|}{35.22}          & \multicolumn{1}{c|}{22.93}          & \multicolumn{1}{c|}{16.56}          & 1.1853               & \multicolumn{1}{c|}{34.36}          & \multicolumn{1}{c|}{21.12}          & \multicolumn{1}{c|}{13.50}        & 1.1258               & \multicolumn{1}{c|}{10.27}          & \multicolumn{1}{c|}{7.57}          & \multicolumn{1}{c|}{10.51}          & 0.3745               \\
ASTGNN                          & \multicolumn{1}{c|}{31.16}          & \multicolumn{1}{c|}{19.26}          & \multicolumn{1}{c|}{12.65}          & 1.0583               & \multicolumn{1}{c|}{32.75}          & \multicolumn{1}{c|}{20.63}          & \multicolumn{1}{c|}{12.47}          & 1.0784               & \multicolumn{1}{c|}{9.68}          & \multicolumn{1}{c|}{7.35}          & \multicolumn{1}{c|}{10.23}          & 0.3569               \\
PDFormer                        & \multicolumn{1}{c|}{29.97}          & \multicolumn{1}{c|}{18.32}          & \multicolumn{1}{c|}{12.10}          & 1.0217               & \multicolumn{1}{c|}{30.18}          & \multicolumn{1}{c|}{19.13}          & \multicolumn{1}{c|}{11.92}          & 1.0205               & \multicolumn{1}{c|}{9.46}          & \multicolumn{1}{c|}{7.12}          & \multicolumn{1}{c|}{10.06}          & 0.3375               \\ 
STAEformer                      & \multicolumn{1}{c|}{30.18}          & \multicolumn{1}{c|}{18.22}          & \multicolumn{1}{c|}{\textbf{11.98}} & 0.9678               & \multicolumn{1}{c|}{29.94}          & \multicolumn{1}{c|}{18.85}          & \multicolumn{1}{c|}{12.03}          & 0.9728               & \multicolumn{1}{c|}{8.69}         & \multicolumn{1}{c|}{6.93}          & \multicolumn{1}{c|}{9.89}          & 0.3083               \\
MultiSPANS                      & \multicolumn{1}{c|}{30.46}          & \multicolumn{1}{c|}{19.07}          & \multicolumn{1}{c|}{13.29}          & 0.9851               & \multicolumn{1}{c|}{30.31}          & \multicolumn{1}{c|}{18.97}          & \multicolumn{1}{c|}{11.85}          & 0.9931               & \multicolumn{1}{c|}{8.57}         & \multicolumn{1}{c|}{6.84}          & \multicolumn{1}{c|}{10.05}          & 0.3149               \\
STG-Mamba                       & \multicolumn{1}{c|}{\textbf{29.53}} & \multicolumn{1}{c|}{\textbf{18.09}} & \multicolumn{1}{c|}{12.11}          & \textbf{0.9218}      & \multicolumn{1}{c|}{\textbf{29.23}} & \multicolumn{1}{c|}{\textbf{18.26}} & \multicolumn{1}{c|}{\textbf{11.59}} & \textbf{0.9271}      & \multicolumn{1}{c|}{\textbf{8.05}} & \multicolumn{1}{c|}{\textbf{6.37}} & \multicolumn{1}{c|}{\textbf{9.64}} & \textbf{0.2648}      \\ \hline
\end{tabular}
}
\end{table}

\subsection{Robustness Analysis}

\begin{table}[htb]
\centering
\caption{Robustness analysis. We choose rush/non-rush, weekend/non-weekend traffic scenarios.}
\label{tab4}
\scriptsize
\begin{tabular}{l|lll|lll}
\hline
\multicolumn{1}{c|}{\multirow{2}{*}{\textbf{Models}}} & \multicolumn{3}{c|}{\textbf{Rush Hours}}                                                                    & \multicolumn{3}{c}{\textbf{Non Rush Hours}}                                                              \\ \cline{2-7} 
\multicolumn{1}{c|}{}                                 & \multicolumn{1}{c|}{\textbf{RMSE}} & \multicolumn{1}{c|}{\textbf{MAE}} & \multicolumn{1}{c|}{\textbf{MAPE}} & \multicolumn{1}{c|}{\textbf{RMSE}} & \multicolumn{1}{c|}{\textbf{MAE}} & \multicolumn{1}{c}{\textbf{MAPE}} \\ \hline
ASTGNN                                                 & \multicolumn{1}{l|}{33.57}         & \multicolumn{1}{l|}{21.05}        & 13.81                              & \multicolumn{1}{l|}{29.76}         & \multicolumn{1}{l|}{18.22}        & 12.49                              \\
PDFormer                                               & \multicolumn{1}{l|}{31.69}         & \multicolumn{1}{l|}{19.54}        & 12.38                              & \multicolumn{1}{l|}{\textcolor{red}{28.95}}         & \multicolumn{1}{l|}{18.16}        & 11.97                              \\
STAEformer                                             & \multicolumn{1}{l|}{31.36}         & \multicolumn{1}{l|}{19.10}        & 12.22                              & \multicolumn{1}{l|}{29.47}         & \multicolumn{1}{l|}{17.98}        & 11.93                              \\
MultiSPAN                                              & \multicolumn{1}{l|}{30.95}         & \multicolumn{1}{l|}{19.28}        & 13.31                              & \multicolumn{1}{l|}{30.04}         & \multicolumn{1}{l|}{18.37}        & 12.54                              \\
STG-Mamba        & \multicolumn{1}{l|}{\textcolor{red}{29.58}}   & \multicolumn{1}{l|}{\textcolor{red}{18.13}}   & {\textcolor{red}{12.15}}      & \multicolumn{1}{l|}{29.23}     & \multicolumn{1}{l|}{\textcolor{red}{18.04}}       & {\textcolor{red}{11.91}}   \\ \hline
\multicolumn{1}{c|}{\multirow{2}{*}{\textbf{Models}}} & \multicolumn{3}{c|}{\textbf{Weekend}}                                                                       & \multicolumn{3}{c}{\textbf{Non Weekend}}                                                                 \\ \cline{2-7} 
\multicolumn{1}{c|}{}                                 & \multicolumn{1}{c|}{\textbf{RMSE}} & \multicolumn{1}{c|}{\textbf{MAE}} & \multicolumn{1}{c|}{\textbf{MAPE}} & \multicolumn{1}{c|}{\textbf{RMSE}} & \multicolumn{1}{c|}{\textbf{MAE}} & \multicolumn{1}{c}{\textbf{MAPE}} \\ \hline
ASTGNN                                                 & \multicolumn{1}{l|}{30.66}         & \multicolumn{1}{l|}{19.08}        & 13.35                              & \multicolumn{1}{l|}{31.82}         & \multicolumn{1}{l|}{19.53}        & 12.28                              \\
PDFormer                                               & \multicolumn{1}{l|}{29.87}         & \multicolumn{1}{l|}{18.26}        & 12.04                              & \multicolumn{1}{l|}{30.39}         & \multicolumn{1}{l|}{18.68}        & 12.23                              \\ 
STAEformer                                             & \multicolumn{1}{l|}{29.83}         & \multicolumn{1}{l|}{18.19}        & 11.98                              & \multicolumn{1}{l|}{30.26}         & \multicolumn{1}{l|}{18.42}        & 12.27                              \\
MultiSPAN                                              & \multicolumn{1}{l|}{30.13}         & \multicolumn{1}{l|}{18.42}        & 12.73                              & \multicolumn{1}{l|}{30.52}         & \multicolumn{1}{l|}{19.31}        & 13.34                              \\
STG-Mamba   & \multicolumn{1}{l|}{\textcolor{red}{29.31}}  & \multicolumn{1}{l|}{\textcolor{red}{18.05}}    & \textcolor{red}{11.92}   & \multicolumn{1}{l|}{\textcolor{red}{29.57}}         & \multicolumn{1}{l|}{\textcolor{red}{18.15}}        & \textcolor{red}{12.16}                        \\ \hline
\end{tabular}
\end{table}

STG data exhibit notable periodicity and diversity, with urban mobility/traffic data showing clear differences between peak travel times in the morning/evening and non-peak periods, as well as between weekdays and weekends. Given these variations due to external environmental changes, it is significant to determine whether deep learning models can effectively model the spatiotemporal dependencies under different conditions. As such, we establish four distinct external scenarios: (a) rush hours on weekdays from 8:00-11:00 AM and 4:00-7:00 PM; (b) non-rush hours during weekdays; (c) weekends (all hours); and (d) non-weekend (all hours). Extensive experiments were conducted on the PeMS04 dataset. Table~\ref{tab4} presents the forecasting results of the four traffic scenarios.

An ideal deep learning-based STG system should be robust to external disturbances, maintaining consistent performance across different scenarios. Compared methods like ASTGNN, PDFormer, and STAEformer show significant performance differences between Rush hour and Non-rush Hour scenarios. Even in the Weekend/Non-Weekend scenario, STG-Mamba demonstrates the smallest performance difference, showing maximum robustness and achieving the best RMSE/MAE/MAPE metrics. In summary, STG-Mamba exhibits superior robustness and outperforms existing baselines in diverse traffic conditions.

%An ideal deep learning-based STG system should maintain robustness to external disturbances (changes of external environment or events). This is specifically reflected in the fact that when facing different scenarios, the model performance should maintain as little difference as possible, rather than being good at some scenarios but being bad at other occasions. Among the compared methods, ASTGNN, PDFormer, and STAEformer exhibit relatively large differences in the prediction results between the Rush hour and Non-rush Hour scenarios. In the Weekend/Non-Weekend scenario, although the performance differences are not as significant as in the Rush hour/Non-Rush hour case, we observe that STG-Mamba still shows the smallest performance difference (the maximum robustness) and achieves the best RMSE/MAE/MAPE metrics. In summary, STG-Mamba proves the best robustness and outperforms existing baselines even facing complex and diversity traffic scenes.

Statistical-based evaluation in spatiotemporal forecasting is indispensible as they provide a quantifiable measure of how well a model captures and predicts complex data dynamics over time and space. Specifically, $R^2$ and $\Delta_{\text{VAR}}$ help in assessing the accuracy and robustness of models under varying conditions, ensuring the reliability of predictions and effectively inform decision-making processes. The Statistical evaluation results are presented in Table~\ref{tab3}. Here, STG-Mamba demonstrates superior performance across datasets, achieving the highest $R^2$ values and the lowest $\Delta_{\text{VAR}}$ scores, indicating the advancement of accuracy and efficiency in handling spatiotemporal dependencies. 

\begin{table}[!htb]
\centering
\scriptsize
\caption{Statistical evaluation results.}
\label{tab3}
\begin{tabular}{c|cc|cc|cc}
\hline
\multirow{2}{*}{\textbf{Model}} & \multicolumn{2}{c|}{\textbf{PeMS04 (Flow)}}                                & \multicolumn{2}{c|}{\textbf{HZMetro}}                                      & \multicolumn{2}{c}{\textbf{KnowAir}}                                      \\ \cline{2-7} 
                                & \multicolumn{1}{c|}{$\Delta_{\text{VAR}}$} & \rm{$R^2$} & \multicolumn{1}{c|}{$\Delta_{\text{VAR}}$} & \rm{$R^2$} & \multicolumn{1}{c|}{$\Delta_{\text{VAR}}$} & \rm{$R^2$} \\ \hline
STGCN                           & \multicolumn{1}{c|}{0.23145}                        & 0.77069              & \multicolumn{1}{c|}{0.25769}                        & 0.74385              & \multicolumn{1}{c|}{0.49851}                        & 0.50487              \\
STSGCN                          & \multicolumn{1}{c|}{0.20538}                        & 0.79456              & \multicolumn{1}{c|}{0.21474}                        & 0.78862              & \multicolumn{1}{c|}{0.47536}                        & 0.52831              \\ 
STG-NCDE                        & \multicolumn{1}{c|}{0.18275}                        & 0.81973              & \multicolumn{1}{c|}{0.19827}                        & 0.80492              & \multicolumn{1}{c|}{0.46173}                        & 0.54125              \\
DDGCRN                          & \multicolumn{1}{c|}{0.13291}                        & 0.87355              & \multicolumn{1}{c|}{0.14796}                        & 0.85511              & \multicolumn{1}{c|}{0.43185}                        & 0.56506              \\
ASTGCN                          & \multicolumn{1}{c|}{0.17289}                        & 0.82698              & \multicolumn{1}{c|}{0.18131}                        & 0.82134              & \multicolumn{1}{c|}{0.42729}                        & 0.58461              \\
ASTGNN                          & \multicolumn{1}{c|}{0.14863}                        & 0.84850              & \multicolumn{1}{c|}{0.15765}                        & 0.84627              & \multicolumn{1}{c|}{0.39726}                        & 0.60563              \\
PDFormer                        & \multicolumn{1}{c|}{0.11849}                        & 0.88190              & \multicolumn{1}{c|}{0.13259}                        & 0.87095              & \multicolumn{1}{c|}{0.36182}                        & 0.64719              \\
STAEformer                      & \multicolumn{1}{c|}{0.09631}                        & 0.90632              & \multicolumn{1}{c|}{0.10383}                        & 0.89738              & \multicolumn{1}{c|}{0.33215}                        & 0.66085              \\
MultiSPANS                      & \multicolumn{1}{c|}{0.06218}                        & 0.93804              & \multicolumn{1}{c|}{0.07596}                        & 0.92561              & \multicolumn{1}{c|}{0.32569}                        & 0.67122              \\ \hline
STG-Mamba                       & \multicolumn{1}{c|}{\textbf{0.04387}}               & \textbf{0.95489}     & \multicolumn{1}{c|}{\textbf{0.05148}}               & \textbf{0.94752}     & \multicolumn{1}{c|}{\textbf{0.30182}}               & \textbf{0.69537}     \\ \hline
\end{tabular}
\end{table}

%时空数据具有显著的周期性和多样性。对于城市交通出行数据来说，其每天早晚高峰出行时段和非高峰时间段的交通数据特征具有明显区别。此外，工作日每天24小时的交通模式和周末两天的交通模式也有很大的不同。面对由于外部环境变化带来的数据特征的显著差异，衡量深度学习模型是否能准确有效地建模对应各种外部环境条件下时空特征是非常有必要的。对此，我们设置了四种不同的外部环境场景，分别为: (a) Traffic rush hours: 周一至周五上午8:00-11:00,下午16:00-19:00; (b) Non rush hours: 周一至周五的剩余时间段. (c) Weekend: 周六和周日的所有时间; (d) Non weekend: 周一至周五的所有时间段,并在PeMS04数据集上进行了广泛的实验。Table 4展示了对应这四种交通场景的时空预测实验结果。我们对比了本文提出STG-Mamba和其他4个表现优越的baseline methods的预测表现，并report了RMSE/MAE/MAPE误差。

%一个理想的基于深度学习构造的STG System应该对于外界的扰动(不同的外部环境变化)保持鲁棒性。这具体体现在面对不同的场景时，其预测效果应该尽可能保持较小的差异性，而不是时好时坏。如Table 4所示，在这些对比的模型中，ASTGNN, PDFormer, STAEformer在Rush Hour场景和Non Rush Hour场景下预测结果具有较大的差异性。在Weekend/Non Weekend场景中，虽然各模型的预测表现差异均没有Rush hour/Non rush hour场景下那样显著，但是我们观察到STG-Mamba仍然表现出最小的差异性(最大的鲁棒性)，并且取得了最佳的RMSE/MAE/MAPE指标。综上所述，我们可以发现STG-Mamba模型具有最强的鲁棒性，在面对复杂多变的交通场景时仍然表现出超越现有基线模型的表现。

\subsection{Ablation Study}

\begin{figure*}[htb]
  \centering
  \includegraphics[width=0.98\columnwidth]{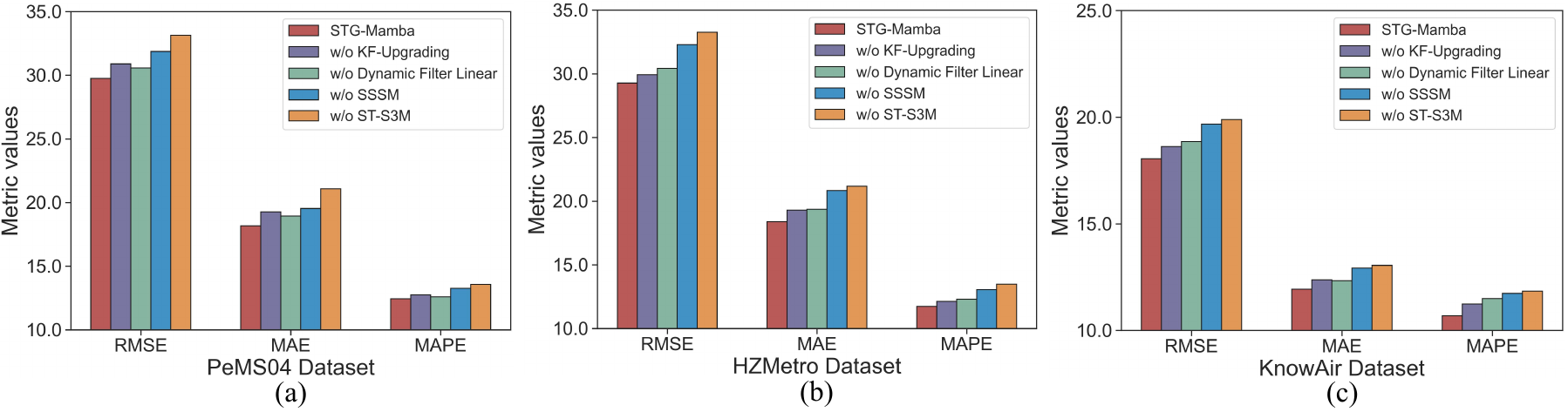}
  \caption{Model component analysis of STG-Mamba.}
  \label{fig5}
\end{figure*}
To investigate the effectiveness of each model component within STG-Mamba, we further design five kinds of model variants, and evaluate their forecasting performance on PeMS04/HZMetro/KnowAir dataset: (I) \textbf{STG-Mamba}: The full STG-Mamba model without any modification. (II) \textbf{STG-Mamba w/o KF-Upgrading}: We replace the KF-Upgrading module in KFGN with the simple sum average operation to fuse the three temporal branches. 
(III) \textbf{STG-Mamba w/o Dynamic Filter Linear}: We replace the proposed Dynamic Filter Linear in KFGN with the basic static graph convolution. (IV) \textbf{STG-Mamba w/o GSSSM}: The Graph State Space Selection Mechansim (GSSSM) is replaced by the basic SSSM~\cite{gu2023mamba}. (V) \textbf{STG-Mamba w/o ST-S3M}: The whole ST-S3M module is removed from GS3B Encoder. 

As illustrated in Figure~\ref{fig5}, STG-Mamba w/o KF-Upgrading shows worse performance than the full model, demonstrating the necessity and suitability of employing statistical learning-based GNN state space upgrading and optimization for SSSM-based methods. The decrease of performance due to the absence of Dynamic Filter Linear demonstrate its effectiveness, and the necessity of designing a suitable adaptive GNN for STG feature learning. Furthermore, removing the State Space Selection Mechansim in ST-S3M results in a substantial degradation of model capability, proving the feasibility of using SSSM as the substitute for Attention mechanism. Finally, the removal of ST-S3M makes the STG-Mamba degrade to a plain GNN-based model, resulting in the lowest performance. 

\section{Conclusions}
In this work, we for the first time introduce deep learning-based selective state space models (SSSM) for spatial-temporal graph learning tasks. We propose STG-Mamba that leverages modern SSSM for accurate and efficient STG forecasting. In STG-Mamba, the ST-S3M module facilitates input-dependent graph evolution and feature selection, successfully integrates STG network with SSSMs. Through the Kalman Filtering Graph Neural Networks (KFGN), the learned STG embeddings achieve a smooth optimization \& upgrading based on statistical learning, aligning with the whole STG selective state space modeling process. Compared with Attention-based methods, STG-Mamba achieves linear-time complexity and substantial reduction in FLOPs and inference time. Extensive empirical studies are conducted on three open-sourced STG datasets, demonstrating the consistent superiority of STG-Mamba over other benchmark methods. We believe that STG-Mamba presents a brand new promising approach to general STG learning fields, offering competitive model performance and input-dependent contextual feature selection under an affordable computational cost.

%We conduct extensive empirical studies on three open-sourced STG datasets from urban traffic, metro In/Out flow, and weather records. Experimental results demonstrate the consistent superiority of STG-Mamba over other benchmark methods in both forecasting performance and computational efficiency. We believe that STG-Mamba presents a brand new promising approach to general STG learning fields, offering competitive model performance and input-dependent context-aware feature selection under an affordable computational cost. 

\bibliographystyle{unsrtnat}
\bibliography{neurips2024}

%%%%%%%%%%%%%%%%%%%%%%%%%%%%%%%%%%%%%%%%%%%%%%%%%%%%%%%%%%%%Appendix%%%%%%%%%%%%%%%%%%%%%%%%%%%%%%%%%%%%%%%%%%%%%%%%%%%%%
\newpage
\appendix
\section{ST-S3M Graph Selective Scan Algorithm \& Parameter Calculation and Upgrade Algorithm}\label{appendA}

%%%%%%%%Graph Selective Scan Algorithm%%%%%%%%%%%%%
\begin{algorithm}[!htb]
\caption{Graph Selective Scan Algorithm}
\label{alg:alg1}
\textbf{Input:} $u,\Delta^{*},\bold{A},\bold{B},\bold{C},\bold{D}$; \\
\textbf{Output:} $z$; 
\begin{algorithmic}[1]
\STATE Obtain the dim of $u$ as $(b, l, d_{\text{in}})$ the second dim of $\bold{A}$ as $n$; \\
\STATE Get the dynamic graph adjacency matrix from DynamicFilter\-GNN: \\
       $\alpha_t \gets \text{DynamicFilter\-GNN.get\_transformed\_adjacency()}$; \\
\STATE Initialize a padding matrix $\text{adj\_padded} \in \mathbb{R}^{d_{\text{in}}\times d_{\text{in}}}$: $\text{adj\_padded} \gets \mathbf{1}^{d_{\text{in}} \times d_{\text{in}}}$; \\
\STATE Filling the padding matrix with Graph information: \\
       $\text{adj\_padded}[:\alpha_t \text{.size}(0), :\alpha_t \text{.size}(1)] \gets \alpha_t$; \\
\STATE Integrate $\Delta^{*}$ and $\text{adj\_padded}$ through multiplication: \\
       $\Delta^{'} \gets \text{matmul}(\Delta^{*}, \text{adj\_padded})$; \\
\STATE Discretize continuous parameters $\bold{A}$ and $\bold{B}$: \\
\STATE $\text{delta}\bold{A} \gets \exp(\text{einsum}(\Delta^{'},\bold{A}))$, where $\text{delta}\bold{A} \in \mathbb{R}^{b \times l \times d_{\text{in}} \times n}$; \\
\STATE $\text{delta}\bold{B}_u \gets \text{einsum}(\Delta^{'},\bold{B},u)$, where $\text{delta}\bold{B}_u \in \mathbb{R}^{b \times l \times d_{\text{in}} \times n}$; \\
\STATE Initialize State $x$ as zeroes with dimension $\mathbb{R}^{b \times d_{\text{in}} \times n}$;\\
\FOR{$i$ from $1$ to $l$}
    \STATE $x \gets \text{delta}\bold{A}[:, i] \times x + \text{delta}\bold{B}_u[:, i]$;
    \STATE $z \gets \text{einsum}(x,\bold{C}[:, i, :])$;
    \STATE Append $z$ to the output list $z_s$;
\ENDFOR
\STATE $z \gets \text{stack}(z_s)$, where $z \in \mathbb{R}^{b \times l \times d_{\text{in}}}$;\\
\STATE Adding the direct gain $\bold{D}$:\\
\STATE $z \gets z + u \times \bold{D}$\\
\STATE Return $z$.
\end{algorithmic}
\end{algorithm}

\begin{algorithm}[htb]
\caption{State Space Selection Mechanism Parameter Calculation \& Upgrade Algorithm.}
\label{alg:alg2}
\textbf{Input}: $x \in \mathbb{R}^{b\times l \times d_{\text{model}}}$, learnable model params $\bold{A}_{log} \in \mathbb{R}^{d_{\text{in}} \times n}$ and $\bold{D}\in \mathbb{R}^{d_{\text{in}}}$;
\\
\textbf{Output}: $y \in \mathbb{R}^{b\times l\times d_{\text{inner}}}$;
\begin{algorithmic}[1]
\STATE Compute input-independent params $\bold{A}\in \mathbb{R}^{d_{\text{in}} \times n}$ and $\bold{D}\in \mathbb{R}^{d_{\text{in}}}$: \\
\STATE $\bold{A}=-\exp(\bold{A}_{log})$, $\bold{D}=\bold{D}$;
\STATE Process input $x$ with Linear Layer $Input\_proj$: \\
\STATE $x_{\text{dbl}}=Input\_proj(x)$, $x_{\text{dbl}} \in \mathbb{R}^{b\times l \times(dt_{\text{rank}} + 2n)}$;
\STATE Dividing $x_{\text{dbl}}$ in order to get the input-dependent params $\Delta,\bold{B},\bold{C}$, \\
where $\Delta \in \mathbb{R}^{b\times l\times dt_{\text{rank}}}$, $\bold{B},\bold{C} \in \mathbb{R}^{b \times l \times n}$;
\STATE Use Linear Layer $dt\_proj$ and $SoftPlus$ activation to adjust $\Delta$: \\
\STATE $\Delta^* = SoftPlus(dt\_proj(\Delta))$, where $\Delta^* \in \mathbb{R}^{b \times l\times d_{\text{in}}}$;\\
\STATE Employ the $\text{GraphSelectiveScan}$ algorithm to proceed with $x,\Delta^{*},\bold{A},\bold{B},\bold{C},\bold{D}$: \\
\STATE $y=\text{GraphSelectiveScan}(x,\Delta^{*},\bold{A},\bold{B},\bold{C},\bold{D})$;
\STATE Return the Output $y$.
\end{algorithmic}
\end{algorithm}

\section{Computational Efficiency Analysis and Comparison}\label{appendB}
\subsection{Theoretical Computational Efficiency Analysis}
Leveraging a data-specific selection process, the increasing of sequence length $L$ or graph network node scale $N$ by $k$-fold can results in huge expansion of model parameters for Transformer-based methods. STG-Mamba introduces a sophisticated algorithm that is conscious of hardware capabilities, exploiting the layered structure of GPU memory to minimize the ensuing cost. In detail, for a batch size of $B$, STG feature dimension of $d'$, and network node number of $N$, STG-Mamba reads the $O(BLd'+Nd')$ input data of $\bold{A}, \bold{B}, \bold{C}, \Delta$ from High Bandwidth Memory (HBM), processes the intermediate stages sized $O(BLd'N)$ in Static Random-Access Memory (SRAM), and dispatches the ultimate output, sized $O(BLd')$, back to HBM. In contrast, if a Transformer-based model is employed, the HBM will read $O(BL^2d')$ input data, and the computational size remains $O(BL^2d')$ in SRAM. In common practices we often have $BL^2d \gg BLd'+Nd'$, which means SSSM-based approach can effectively decrease the computational overhead in large-scale parallel processing, such as neural network training. Additionally, by not retaining intermediate states, it also diminishes memory usage, since these states are recalculated during the gradient assessment in reverse phase. SSSM's GPU-optimized design ensures STG-Mamba operates with a linear time complexity $O(L)$ relative to the length of the input sequence, marking a substantial speed advantage over traditional transformers' dense attention computations, which exhibit quadratic time complexity $O(L^2)$.

\subsection{Experimental Computational Efficiency Analysis}
STG-Mamba offers substantial improvement in computational efficiency in addition to performance gain. In order to quantatively evaluate computational efficiency, we select the widely recognized Floating Point Operations (FLOPs) and Inference Time as criteria. For both of the two criteria, a smaller value indicates better model performance. STG-Mamba is compare with other Transformer-based benchmark methods: ASTGNN, PDFormer, and STAEformer, on the same testing set. 

Table~\ref{tabA} showcases the inference time comparison on PeMS04 and KnowAir testing set, we report the average inference time of every \textbf{100 batch} with Sequence\_Length=12, Batch\_Size=48. It can be seen that ASTGNN is relatively redundancy, with the lowest inference time among the compared Transformer-based methods. STG-Mamba gains 18.09\% and 20.54\% performance increase in inference time than the 2nd best performance model (STAEformer). 

\begin{figure*}[htb]
  \centering
  \footnotesize
  \includegraphics[width=0.75\columnwidth]{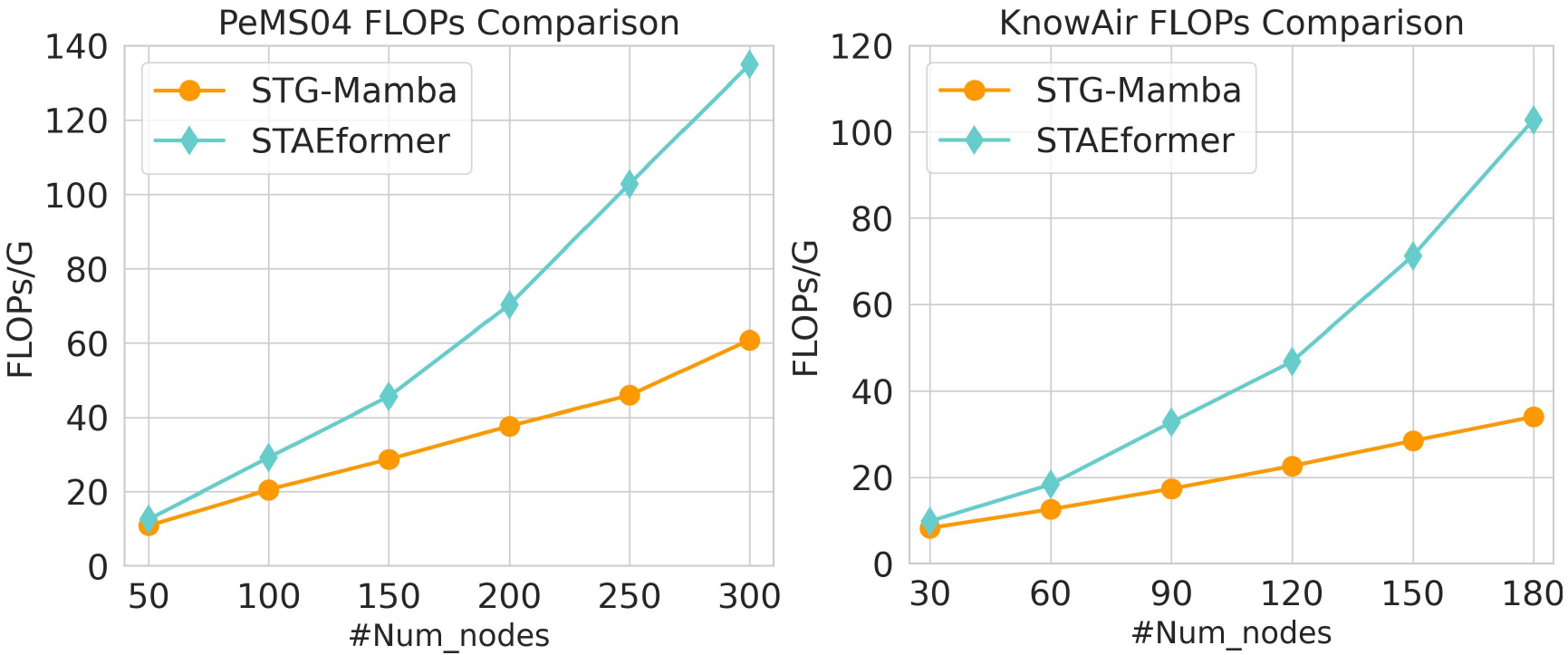}
  \caption{(Computational Efficiency Eval) FLOPs comparison between STG-Mamba and STAEformer on PeMS04/KnowAir dataset under different STG node number settings.}
  \label{fig-appendA}
\end{figure*}

We further compare the computational FLOPs increment with accordance to the scale complexity of STG network. Specifically, we use the number of node individuals to represent the scale of STG network. In the experiments, we select different number of nodes scales: $[50,100,150,200,250,300]$ for PeMS04 and $[30,60,90,120,150,180]$ for KnowAir, respectively. STG-Mamba is compared with the benchmark STAEformer in every settings. Figure~\ref{fig-appendA} visualizes the FLOPs evaluation under different STG scale complexity. We can easily find that STG-Mamba enjoys a linear increment in FLOPs, while STAEformer shows quadratic increment in FLOPs. At the initial small-scale setting, the FLOPs for STG-Mamba/STAEformer are quite similar. However, with the increasing of network complexity and scale, the difference becomes more significant. At the 300 node setting of PeMS04, the FLOPs for STG-Mamba and STAEformer is 60.85G and 134.92G (2.22 times more), respectively. At the 180 node setting of KnowAir, the FLOPs for STG-Mamba and STAEformer is 34.03G and 102.79G (3.54 times more), respectively. 

Benefiting from the linear computational increment brought by modern selective state space models, STG-Mamba and alike models demonstrate stronger capability and computational efficiency for modeling large-scale STG data, which holds promise for addressing the computational overhead in future large-scale STG learning.

\begin{table}[htb]
\centering
\small
\caption{Test inference time comparison with benchmark Transformer-based methods.}
\begin{tabular}{c|c|c}
\hline
Model      & Inference Time on PeMS04 & Inference Time on KnowAir \\ \hline
ASTGNN     & 3.285 s                   & 3.018 s                     \\ %\hline
PDFormer   & 2.431 s                   & 2.357 s                     \\ %\hline
STAEformer & 2.106 s                   & 1.933 s                     \\ \hline
STG-Mamba  & \textbf{1.547 s}          & \textbf{1.481 s}            \\ %\hline
performance $\uparrow$  & \small{+26.54\%}    & \small{+23.38 \%}     \\ \hline
\end{tabular}
\label{tabA}
\end{table}

\section{Evaluation Metrics Calculation Method}\label{appendC}

We employ RMSE/MAE/MAPE/$std_{\text{MAE}}$ as the \textbf{deterministic evaluation metrics}, all the indicators are smaller the better; $\rm{\Delta_{\text{VAR}}}$ (Change in Variance Accounted For) and $\rm{R^2}$ (Coefficient of Determination) as \textbf{statistic-based evaluation metric}, where $\rm{\Delta_{\text{VAR}}}$ is the smaller the better, $\rm{R^2}$ is the larger the better. The reported results in Experiment section are the average value of 10 runs.

Here, $std_{\text{MAE}}$ refers to the standard deviation of multiple MAE values in the testing stage. $std_{\text{MAE}}$ provides a measure of consistency for MAE results across multiple evaluation sets. A smaller $std_{\text{MAE}}$ indicates the model performs more stably under different conditions, with less variation in prediction errors. The calculation formula for $std_{\text{MAE}}$ is as follows: Step1-Calculate the average MAE $\mu$: $\mu = \frac{1}{n} \sum_{i=1}^{n} \text{MAE}_i$. Here, $n$ denotes the number of batches in the testing dataset, with the MAE value computed for each batch. $\text{MAE}_i$ is the MAE result of the $i$-th batch. Step2-Calculate the standard deviation: $\text{std}_{\text{MAE}} = \sqrt{\frac{\sum_{i=1}^{n} (\text{MAE}_i-\mu)^2}{n}}$.

$R^2$ refers to the Coefficient of Determination is a statistical index used for measuring the proportion of variance in the dependent variable that is predictable from the independent variables. $R^2$ provides a gauge of the effectiveness of the model in explaining the variations in the observed data. A higher $R^2$ value indicates the model has a higher predictive accuracy and explains a substantial proportion of the variance in the dependent variable. $R^2$ is formulated as follows:
$R^2 = 1 - \frac{\sum_{i=1}^{n} (y_i - \hat{y}_i)^2}{\sum_{i=1}^{n} (y_i - \bar{y})^2}$ where $y_i$ denotes the ground-truth values, $\hat{y}_i$ denotes the model predictions, $\bar{y}$ is the mean of the ground-truth values. $\sum (y_i - \hat{y}_i)^2$ indicates the unexplained variance by the model, while the denominator, $\sum (y_i - \bar{y})^2$, is the total variance in the observed data.

Change in Variance Accounted For ($\Delta_{VAR}$) quantifies the improvement or regression in how much variance the model can account for compared to a previous models or baselines. This metric is particularly valuable in iterative model optimization where understanding the variance captured by model updates is crucial. $\Delta_{VAR}$ is calculates as follows:
$\Delta_{VAR} = \frac{\text{Var}(y - \hat{y})}{\text{Var}(y)}$. $\text{Var}(y - \hat{y})$ computes the variance of residuals, and $\text{Var}(y)$ is the variance of the ground-truth data. The difference between the variance ratio of the current and previous models provides the change in variance accounted for, reflecting the model's improvement in explaining the data variability.

\section{Formulation of Selective State Space Models for STG Learning}\label{appendD}

SSSMs like structured state space sequential models (S4) and Mamba, initially designed for sequential data, are adapted for Spatial-Temporal Graph (STG) data modeling by leveraging their structure to handle spatiotemporal dependencies dynamically. The adaptation involves discretizing the continuous system described by linear Ordinary Differential Equations (ODEs) into a format amenable to deep neural networks modeling and optimization, enhancing computational efficiency while capturing the evolving dynamics of the STG networks.

Modern selective state space models, such as Mamba, build upon classical continuous systems by incorporating deep learning techniques and discretization methods suitable for complex time-series data. Mamba specifically maps a one-dimensional input sequence $x(t) \in \mathbb{R}$ through intermediate implicit states $h(t) \in \mathbb{R}^N$, culminating in an output $y(t) \in \mathbb{R}$. The fundamental process is described by a modified set of state space equations that include discretization adjustments to accommodate deep learning frameworks:
\begin{equation}
\begin{array}{lr}
h'(t) = \overline{\bold{A}} h(t) + \overline{\bold{B}} x(t)\\
y(t) = \bold{C} h(t)
\label{eq:eq2}
\end{array}
\end{equation}

\noindent where $\overline{\bold{A}} \in \mathbb{R}^{N \times N}$ and $\overline{\bold{B}} \in \mathbb{R}^{N \times 1}$ are discretized representations of the continuous state transition and input matrices, respectively. $\bold{C} \in \mathbb{R}^{1 \times N}$ is the output matrices. The discretization process involves transformation techniques that convert $\bold{A}$ and $\bold{B}$ from their continuous forms to computational efficient formats. These transformations are determined by a selective scan algorithm that utilizes a convolutional approach to model interactions across temporal dimension:

\begin{equation}
\begin{array}{lr}
\overline{\bold{A}} = \exp(\Delta \bold{A})\\
\overline{\bold{B}} = \Delta \bold{B} \cdot (\exp(\Delta \bold{A}) - \bold{I})
\label{eq:eq3}
\end{array}
\end{equation}

Mamba extends classic state space model functionalities by incorporating a selective mechanism that dynamically adapts to the input sequence characteristics. This adaptability is achieved through a novel architecture that selectively applies convolution kernels across the sequence, optimizing the model’s response to spatiotemporal variations:

\begin{equation}
\begin{array}{lr}
\overline{\bold{K}} = (\bold{C}\overline{\bold{B}}, \bold{C}\overline{\bold{A}\bold{B}}, \dots, \bold{C}\overline{\bold{A}^{L-1}\bold{B}})
\label{eq:eq4}
\end{array}
\end{equation}

\noindent where $\overline{\bold{K}}$ denotes a structured convolutional kernel, capturing dependencies across spatial and temporal dimensions, and $L$ represents the length of the input sequence $x$. This selective convolutional approach significantly enhances the model's capability to forecast spatiotemporal data by focusing computation on critical features of the data sequence, thus improving both accuracy and efficiency in predictions.

\section{Discussion of Limitations}\label{appendE}

STG-Mamba for spatial-temporal graph learning demonstrates significant advantages, such as superior forecasting accuracy, model robustness, computational efficiency, and the ability to effectively handle datasets with noise and uncertainty. Despite these promising results, there are some limitations to consider:
\begin{itemize}
\item Although STG-Mamba has been extensively evaluated on three diverse open-sourced datasets (Weather/Traffic/Urban Metro), these datasets may not fully representing the wide range of possible spatiotemporal scenarios, potentially limiting the generalizability of the model. Future work should aim to test the model on a broader range of datasets from different geographical regions and temporal scales.

\item The model interpretability is another concern worthy of further investigation. Same as other deep learning-based methods, the black-box nature of Mamba architecture makes it not easy to interpret how specific predictions are made. Enhancing the interpretability deep learning-based SSSMs could be beneficial, particularly in applications where understanding the decision-making process is crucial.

\item While the novel integration of Kalman Filtering statistic learning-based optimization with dynamic graph neural networks helps in managing noise and uncertainty, the model's performance can still be affected by the quality and completeness of input dataset. Developing robust data pre-processing techniques and improving the model's ability to directly handle incomplete or noisy data would further enhance its reliability.

\end{itemize}

We believe by addressing the aforementioned limitations, the future iterations of STG-Mamba can achieve even greater accuracy, robustness, and applicability across a broader range of spatial-temporal graph learning tasks.

\section{Illustration of the STG feature selection procedure} \label{appendF}
\begin{figure*}[!htb]
  \centering
  \includegraphics[width=0.99\columnwidth]{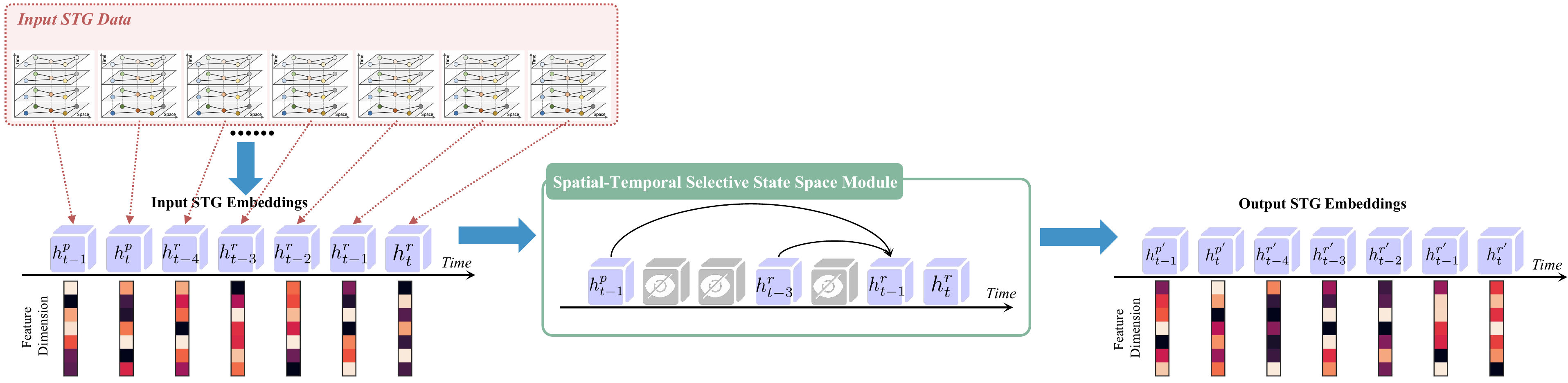}
  \caption{Illustration of the spatial-temporal state space selection mechanism that facilitates input-dependent feature selection and focusing.}
  \label{fig3}
\end{figure*}

Figure~\ref{fig3} is the illustration of the spatial-temporal state space selection mechanism that facilitates input-dependent feature selection and focusing.

\section{Related Works} \label{appendG}

Owning to the strong ability of capturing dynamic features across both temporal and spatial dimension, attention-based methods, such as Spatial-Temporal Transformer and its variants have recently been broadly implemented in spatial-temporal graph (STG) learning tasks. In the domain of STG traffic forecasting, seminal works have emerged that leverage these methods to enhance prediction accuracy. ASTGCN~\cite{guo2019attention} introduces an innovative approach combining spatial-attention \& temporal attention mechanisms with graph neural networks. ADCT-Net~\cite{kong2024adct} presents to incorporate a dual-graphic cross-fusion Transformer for dynamic traffic forecasting.~\cite{huo2023hierarchical} integrates hierarchical graph convolutional networks with Transformer networks to capture sophisticated STG relationships among traffic data. For weather forecasting,~\cite{liang2023airformer} exemplifies how Transformer-based models can be utilized for accurate air quality forecasting across vast geographical areas.~\cite{chen2023group} further demonstrates the power of GNNs in predicting air quality, emphasizing the importance of group-aware mechanisms in enhancing STG forecasting performance. In the realm of Social Recommendation,~\cite{zhang2023ADT} introduces an Adaptive Disentangled Transformer (ADT) framework that optimizes the disentanglement of attention heads in different layers of Transformer architecture.~\cite{wei2023lightgt} explores the improved efficiency of multimedia recommendation by introducing modal-specific embeddings and a lightweight self-attention.

Although Transformer-based methods demonstrate notable improvements in STG relationship learning~\cite{chen2022bidirectional,cong2021spatial}, their modeling ability on large-scale STG networks and Long-Term Sequential Forecasting (LTSF) tasks is greatly hindered by the quadratic computational complexity $O(n^2)$ provided by its core component--attention mechanism~\cite{zhou2021informer,zhou2022fedformer,rampavsek2022recipe}. Moreover, attention mechanisms usually approximate the full-length historical data, or encode all the context features, which may not be necessary for modeling the long-term dependencies of STG learning tasks. As such, the research world has been actively looking for counterparts of Attention-based models.

\end{document}